\documentclass[10pt,journal]{IEEEtran}

\usepackage{graphicx}
\usepackage{epstopdf}
\usepackage{epsfig}
\usepackage{subfigure}
\usepackage{algorithm}
\usepackage{algorithmic}
\usepackage{multirow}
\usepackage{booktabs}
\usepackage[colorlinks,linkcolor=black,anchorcolor=black,citecolor=black]{hyperref}
\usepackage{amssymb}

\ifCLASSINFOpdf

\else

\fi

\begin{document}

\title{\vspace{-0.75em}Semantic Communication with Adaptive Universal Transformer}

\author{Qingyang~Zhou,~Rongpeng Li, Zhifeng Zhao, Chenghui Peng,  and Honggang Zhang\vspace{-2em}}


\maketitle

\begin{abstract}
With the development of deep learning (DL), natural language processing (NLP) makes it possible for us to analyze and understand a large amount of language texts. Accordingly, we can achieve a semantic communication in terms of joint semantic source and channel coding  over a noisy channel with the help of NLP. However, the existing method to realize this goal is to use a fixed transformer of NLP while ignoring the difference of semantic information contained in each sentence. To solve this problem, we propose a new semantic communication system based on Universal Transformer. Compared with the traditional transformer, an adaptive circulation mechanism is introduced in the Universal Transformer. Through the introduction of the circulation mechanism, the new semantic communication system can be more flexible to transmit  sentences with different semantic information, and achieve better end-to-end performance under various channel conditions.
\end{abstract}


\IEEEpeerreviewmaketitle

\section{Introduction}

\IEEEPARstart{W}{ithin} more than 70 years since Shannon established information theory, scholars’ study mainly focused on the first   level of communication about how to accurately and effectively transmit symbols from the transmitter to the receiver. However, in recent years, with the astonishing development of artificial intelligence, natural language processing (NLP) and other supporting technologies, the intelligent  level of communication system and its cognitive ability to the outside world are constantly enhanced, which provides the promising possibility to carry out the second level of semantic communication. Accordingly, semantic communication has been attracting ever-growing interests as a potential trend in the field of next-generation mobile communications.


Compared with the traditional communication, semantic communication points to greater advantages and wider application scenarios. Within the semantic communication environment, the information sender can better understand the purpose of transmission, simplify the transmitted data and eliminate the transmission of redundant information. 
Moreover, according to the prior knowledge and the received context information within the communication process, the receiver can carry out an intelligent error correction as well as an appropriate recovery of the received information, so as to make the transmission of the message more compressive and accurate.

Thanks to the rapid development of deep learning enabled natural language processing techniques, the time is coming to deepen the concrete research on semantic communication. 
Referring to [2]-[7], we first analyze various typical structures of existing semantic communication. We find that the existing semantic communication systems, compared with the traditional communication systems, have really achieved several advantages more or less. However, all of them have unchanged structures in the face of semantic differences in different sentences over changing communication channels and signal-to-noise ratio  (SNR). So, we feel necessary to design a new semantic communication system that can flexibly deal with the semantic differences and adaptively adjust itself according to the changed channel conditions. In the latest two papers \cite{[2]} \cite{[3]}, Xie \emph{et al}. propose a new semantic communication system based on transformer with fixed attention structure, which obtains greatly improved performance in the low SNR region. Inspired by them, we design a novel semantic communication system based on Universal Transformer (UT) by introducing adaptive circulation mechanism that breaks the fixed structure of conventional transformer [12]. Universal Transformer is capable of giving loop play to its own circulation mechanism, and respond to different 
semantic information through different cycles, so that the semantic communication system can be more adaptive in handling different communication  situations.

 

\section{Semantic Communication System Model and Mechanism}
\subsection{Problem Description}
As shown in Fig. 1, taking into account the shared common knowledge of both transmitter and receiver, the semantic transmitter can map a sentence with context information, \textbf{s}, into a complex symbol stream, \textbf{x}, and then transmits it through the physical communication channel such as AWGN (additive white Gaussian noise) channel or Rayleigh fading channel under certain propagation situation. The received symbol stream, \textbf{y}, is decoded at the semantic receiver, and restored to the target sentence through common knowledge. The transmitter and receiver can be designed using deep neural networks (DNNs) jointly, for which NLP enables DNNs to  train an end-to-end model for transmitting information with variable-length sentences in different languages.

For the semantic communication system, we define the inputting sentences, \textbf{s} = [\emph{w}$_{1}$,\emph{w}$_{2}$,\dots,\emph{w}$_{\emph{L}}$], where  \emph{w}$_{\emph{L}}$ represents the \emph{L}-th word in the sentences.  The whole communication system consists of two sections, namely the semantic section and the channel section. The semantic section contains semantic encoder and semantic decoder to extract and recover the semantic information from the transmitted symbols. Similarly, the channel section contains channel encoder and channel decoder to guarantee the robust transmission of semantic symbols over the physical communication channel.  The encoded symbol stream can be represented by
\begin{equation}
\textbf{x} = \emph{C}_{\alpha}(\emph{S}_{\beta}(\textbf{s}))
\end{equation}
where $\textbf{x} \in \mathbb{C}^{L \times K}$ is the complex channel vector for transmission, K is the number of symbols for each word, $\emph{S}_{\beta}(\cdot)$ is the semantic encoder with the parameter set $\beta$ and $\emph{C}_{\alpha}(\cdot)$ is the channel encoder with the parameter set ${\alpha}$. 
If \textbf{x} is sent, the signal received at the receiver will be  
\begin{equation}
\textbf{y} = \emph{h}\textbf{x} + \emph{\textbf{n}}
\end{equation}
where $\textbf{y} \in \mathbb{C}^{L \times K}$ is  the corresponding complex channel output vector, $\emph{h} \in \mathbb{C}$ is the channel gain  that remains constant throughout the transmission, and $\textbf{n} \in \mathbb{C}^{L \times K}$ is independent and identically distributed circularly symmetric complex Gaussian
noise vector with zero mean and variance $\delta^2$. 
When we consider only AWGN channel, the channel will be set as $\textbf{y} = \textbf{x} + \emph{\textbf{n}}$. 

Furthermore as shown in Fig. 1, the transmitted symbols can be restored through the channel decoder and semantic decoder. The finally decoded signal can be represented as 
\begin{equation}
\hat{\textbf{s}} = \emph{S}^{-1}_{\gamma}(\emph{C}^{-1}_{\delta}(\textbf{y}))
\end{equation}
where $\hat{\textbf{s}}$ is the recovered sentence, $\emph{C}^{-1}_{\delta}(\cdot)$ is the channel decoder with the parameter set $\delta$ and $\emph{S}^{-1}_{\gamma}(\cdot)$ is the semantic decoder with the paramter set $\gamma$.

 The optimal goal of the designed semantic communication system is to minimize the semantic errors while reducing the totally necessary number of symbols to be transmitted, in order to keep the linguistic meaning between s and $\hat{\textbf{s}}$ unchanged. We intend to set up the semantic and channel encoding jointly based on a new DNN framework. 
 
 \begin{figure}[t]
	\centering
	\includegraphics[scale=0.29]{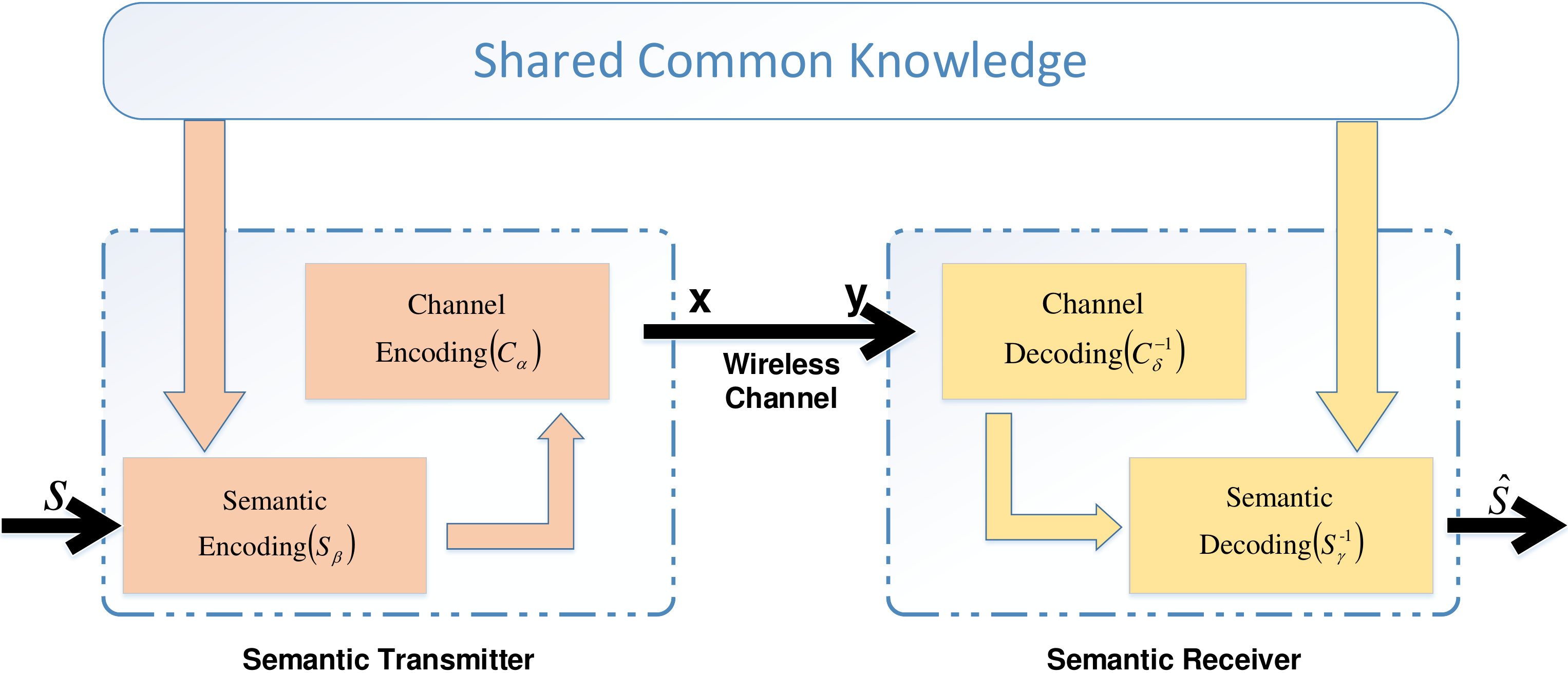}
	\caption{The general framework of semantic communication system.}
	\label{fig:1}       
\end{figure}

\begin{algorithm}[h]
	\caption{ACT halting mechanism} 

	\label{alg:alg1}
 
       
        \textbf{Input:}3D tensor Input \textbf{x}
        
        \textbf{Input:}threshold $0\leq$ T $\leq1$
        
        \textbf{Input:}Maximum number of cycles m
        
        \textbf{Initialization:}Initialize cumulative halting score \textbf{h}, Remainder value \textbf{R}, cumulative number of cycles \textbf{N}
        
        \textbf{Output:}3D tensor \textbf{y}
        
        \textbf{Output:}ponder cost \textbf{$\rho$}
        
       
       
       
       
        1: \textbf{While:} \textbf{h} \textless T and \textbf{N} \textless m do 
       
        2: Compute current step halting score \textbf{p} = $\emph{f}_{\varphi}(\textbf{x})$
       
        3: Mask for Inputs which have not halted \textbf{A} = \textbf{p}, set elements less than 1 in \textbf{A} to 1, elements greater than 1 to 0
      
        4: Mask for Inputs which halted at this step \textbf{B} = \textbf{h}+\textbf{p}$\bigodot$\textbf{A}, set elements less than T in \textbf{B} to 0, elements greater than T to 1
        
        5: Mask for inputs which haven't halted, and didn't halt this step 
        \textbf{A} = \textbf{h}+\textbf{p}$\bigodot$\textbf{A}, set elements less than T in \textbf{A} to 1 and elements greater than T to 0
       
        6: Add the halting probability for this step to the halting \textbf{h} = \textbf{h}+\textbf{p}$\bigodot$\textbf{A}
       
        7: Compute remainders for the inputs which halted at this step \textbf{R} = \textbf{R}+\textbf{B}$\bigodot$(1-\textbf{h})
       
        8: Add the remainders to those inputs which halted at this step \textbf{h} = \textbf{h}+\textbf{B}$\bigodot$\textbf{R}
       
        9: Update the number of cycles \textbf{N} = \textbf{N}+\textbf{A}+\textbf{B}
       
       10: Compute the weight to be applied to the output \textbf{W} = \textbf{p}$\bigodot$\textbf{A}+\textbf{R}$\bigodot$\textbf{B}
       
       11: Apply the Block on the Input \textbf{x} = $\emph{F}(\textbf{x})$
       
       12: Update output \textbf{y} = \textbf{y}$\bigodot$\textbf{(1-W)}+\textbf{x}$\bigodot$\textbf{W}
       
       13: \textbf{End while}
       
       14: \textbf{$\rho$} = \textbf{R} + \textbf{N}
       
       15: Return \textbf{y},\textbf{$\rho$}

\end{algorithm}

\subsection{Proposed Model and Mechanism}
Compared with the existing semantic communication approaches, we take advantage of UT [12] to extract the semantic features from the texts to be transmitted. Different from the recently booming transformer with fixed number of multiple layers, UT introduces circulation mechanism in the structure of transformer. In sequential sentence processing system, certain sentences are usually more ambiguous than the others. What’s more, in the process of information transmission, noise and interference will also be received. 
It is therefore reasonable to allocate more computing resources to those  sentences with more complex semantic information and  the corresponding symbols that are seriously disturbed by noise. On the other hand, Adaptive Computation Time (ACT)  [13] is a mechanism for dynamically modulating the number of computational steps needed to process each input symbol, which are based on a scalar halting probability predicted by the model at each step.


\begin{figure*}[t]
	\centering
	\includegraphics[scale=0.293]{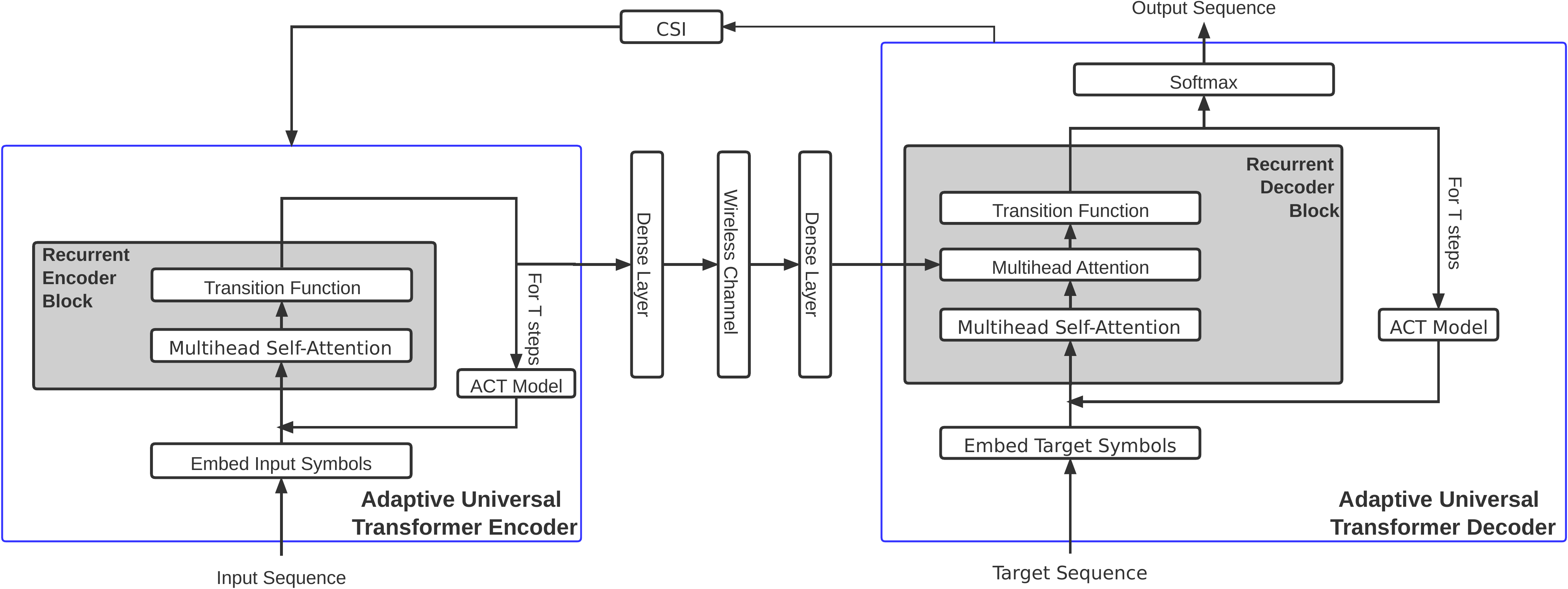}
	\caption{The proposed adaptive UT structure for the semantic communication system.}
	\label{fig:2}       
\end{figure*}

The concrete ACT halting mechanism is described in Algorithm 1. Firstly, by a dense layer with output dimension 1, $\emph{f}_{\varphi}(\cdot)$, we can get the halting score of the current input vector, \textbf{p}. According to the halting score, we can calculate the Remainder value, \textbf{R}, mask for the input which is still running and has stopped, \textbf{A} and \textbf{B}.
Then make the input update through the block, $\emph{F}(\cdot)$. Repeat the above steps until the cumulative halting score, \textbf{h}, exceeds the threshold or the number of cycles \textbf{N} reaches the maximum number preset. Finally return the output result of the block and the ponder cost, \textbf{$\rho$}, which is consisted of \textbf{R} and \textbf{N}. More details about ACT can be found in [13]. After adding ACT halting mechanism, the loss function of the whole system can be written as
\begin{equation}
\emph{L}_{total} = \emph{L}_{origin}+\textbf{$\rho$}
\end{equation}
Based on the original loss function, $\emph{L}_{origin}$, the gradient of {$\rho$} is back propagated at the same time. By training the original blcok and act model together, the whole system can learn to choose the appropriate time to exit the cycle in the face of different situations without affecting the system performance.

\begin{algorithm}[ht]
	\caption{Train the whole semantic communication system } 

	\label{alg:alg2}
 
       1: \textbf{Initialization:}Initialize the weights \textbf{W} and bias b
       
       2: \textbf{Input:}The transmitted sentence \textbf{s}
 
       3: \textbf{Transmitter:}
       
       \quad \ Semantic encoder : $\emph{S}_{\beta}(\textbf{s})$
       $\rightarrow \textbf{M}$
       
              \quad \ Channel encoder : $\emph{C}_{\alpha}(\textbf{M})$
       $\rightarrow \textbf{x}$
       
       4: Transmit \textbf{x} over the channel
       
       5: \textbf{Receiver:}

       \quad \ Channel decoder : $\emph{C}_{\sigma}(\textbf{y})$
       $\rightarrow \hat{\textbf{M}}$

              \quad \ Semantic decoder : $\emph{S}_{\gamma}(\hat{\textbf{M}})$
       $\rightarrow \hat{\textbf{s}}$
       
       6: Compute loss function $\emph{L}$.
       
       7: Train $\alpha,\beta,\gamma,\sigma$ by Gradient descent
       
       8: \textbf{Output:}The whole network $\emph{S}_{\beta}(\cdot)$,
       $\emph{C}_{\alpha}(\cdot)$,
       $\emph{C}_{\sigma}^{-1}(\cdot)$,
       $\emph{S}_{\gamma}^{-1}(\cdot)$

\end{algorithm}

Accordingly, the new semantic communication  system is built up as illustrated in Fig 2. In concrete, the transmitter consists of a semantic encoder to extract the semantic features and a channel encoder to generate the symbols to facilitate the transmission subsequently. The semantic encoder includes Universal Transformer encoder and the channel encoder using dense layers with different units. The propagation channel is interpreted as one neural layer in the model. Similarly, the receiver is composited with a channel decoder for symbol detection and a semantic decoder for text estimation and recovery. The channel decoder includes dense layers with different units and the semantic decoder includes Universal Transformer decoder.

Cross-entropy (CE) is used as the loss function to measure the difference between \textbf{s} and $\hat{\textbf{s}}$, which can be formulated as     \qquad \qquad \qquad \quad
     $\emph{L}_{CE}(\textbf{s},\hat{\textbf{s}},\alpha,\beta,\gamma,
\delta) =  $
\begin{equation}
-\sum_{l=1}
(\emph{q}(\emph{$w_{l}$})log(\emph{p}(\emph{$w_{l}$}))+
(1-\emph{q}(\emph{$w_{l}$}))log(1-\emph{p}(\emph{$w_{l}$})))
\end{equation}
where $\emph{q}(\emph{$w_{l}$})$ is the probability that the $\emph{l}$-th word, $\emph{$w_{l}$}$, appears in the transmitted sentence \textbf{s}, and $\emph{p}(\emph{$w_{i}$})$ is the predicted probability that the $\emph{i}$-th word, $\emph{$w_{i}$}$, appears in the estimated sentence $\hat{\textbf{s}}$. Since the CE can measure the difference between the sentence $\textbf{s}$ and $\hat{\textbf{s}}$, through reducing the loss value of CE, semantic communication system can learn to recover the transmitted information  influenced by the noise and interference.
The final loss function can be formulated as
\begin{equation}
\emph{L}_{total} = \emph{L}_{CE}+\textbf{$\rho$}
\end{equation}
By reducing the value of the loss function, $\emph{L}_{total}$, the whole system learns to use the appropriate number of cycles while ensuring the accuracy of information transmission.

The training process of the whole system is given in  Alogorithm 2. 
First,  initialize the weights \textbf{W} and bias b of the whole  neural networks. Then we input the sentence, \textbf{S}, through the embedding layer to get the vector representation of the sentence.
 Through the semantic encoder, the semantic coding result, \textbf{M},
 is determined by the semantic features of the sentence. Then, \textbf{M} is encoded into symbols \textbf{x} via the channel encoder to deal with the noise from the physical channel. After passing through the physical channel, the receiver receives the distorted symbols, \textbf{y}, which are further decoded by the channel decoder layer. Afterwards, the transmitted sentences are restored by the semantic decoder layer. Finally, in order to get a better transmission result, we compare the difference between the transmitted sentence, \textbf{s}, and the recovered sentence  $\hat{\textbf{s}}$, and optimize the whole system by the stochastic gradient descent (SGD) approach according to the loss function $\emph{L}_{total}$.

\section{Numerical Results}
In this Section, we compare the proposed method with the existing semantic communication algorithms and the traditional joint source coding and channel coding approaches under the AWGN channel and Rayleigh fading channel respectively, where we assume channel state information (CSI) is available from the feedback ACK signal for all considered schemes.

\begin{figure*}[ht]
\centering
\begin{minipage}{0.32\textwidth}
\includegraphics[width=5.5cm,height=3.8cm]{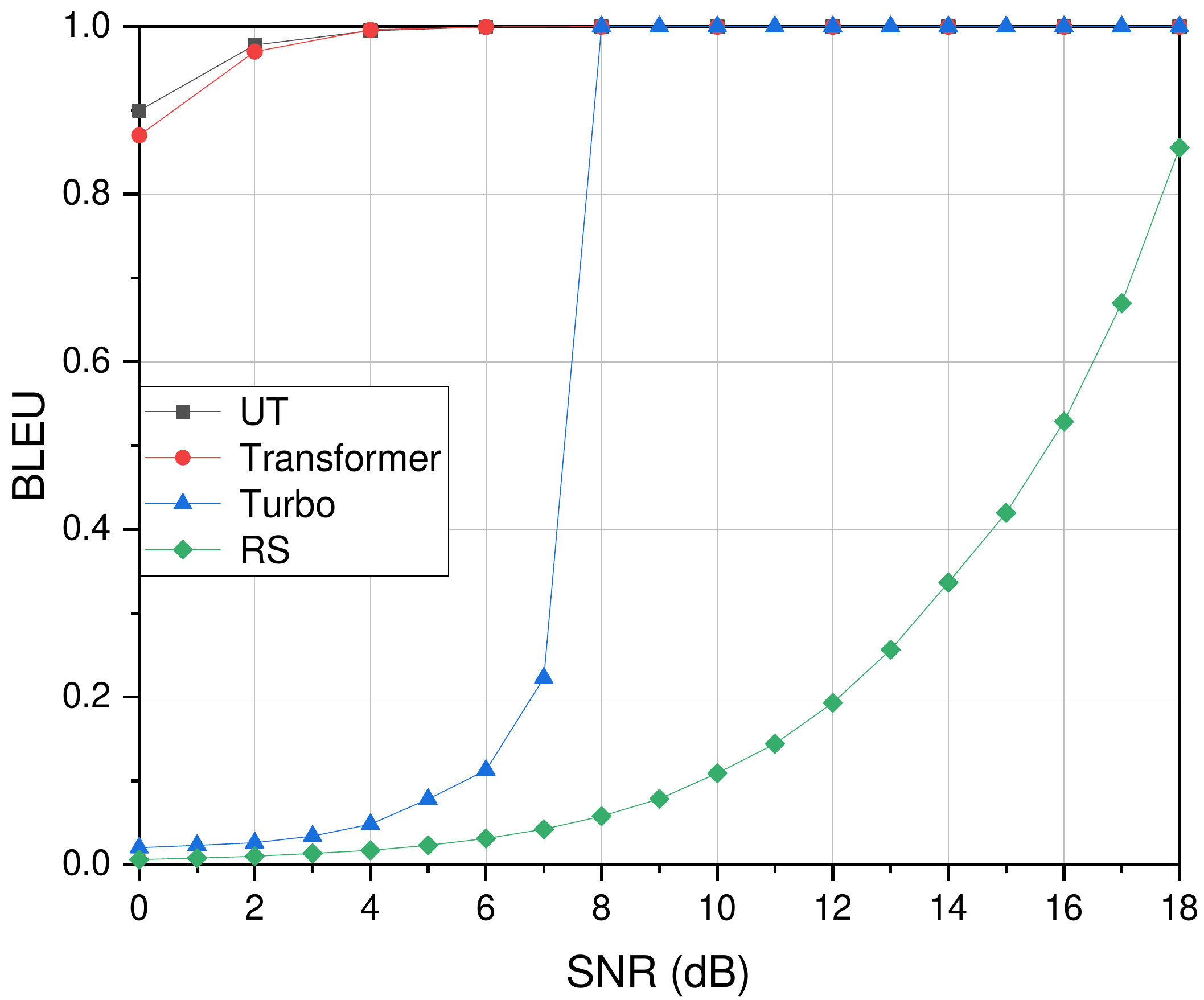}
\caption{BLEU score versus SNR in AWGN channel.}
\end{minipage}
\begin{minipage}{0.32\textwidth}
\includegraphics[width=5.5cm,height=3.8cm]{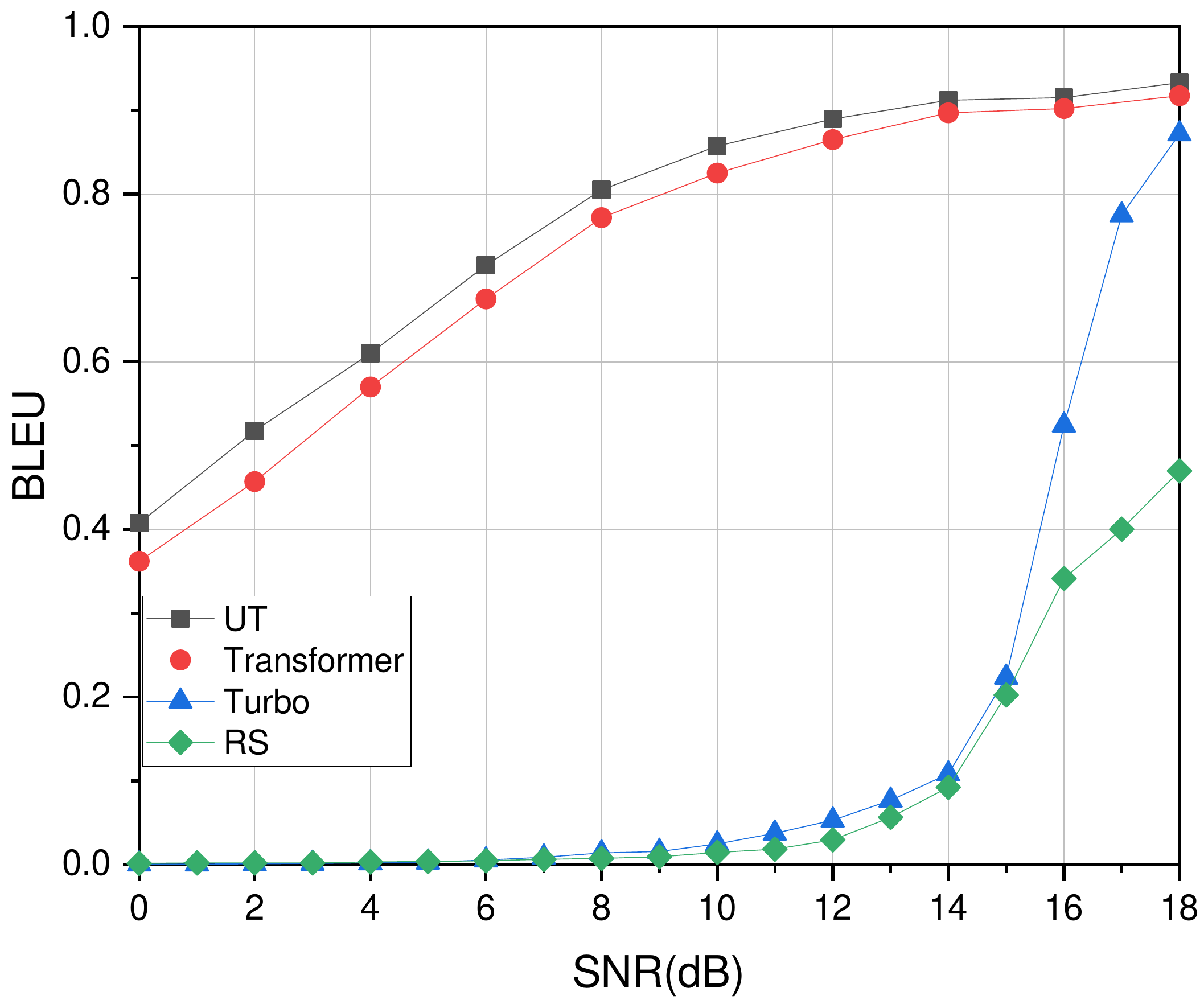}
\caption{BLEU score versus SNR in Rayleigh fading channel.}
\end{minipage}
\begin{minipage}{0.32\textwidth}
\includegraphics[width=5.5cm,height=3.8cm]{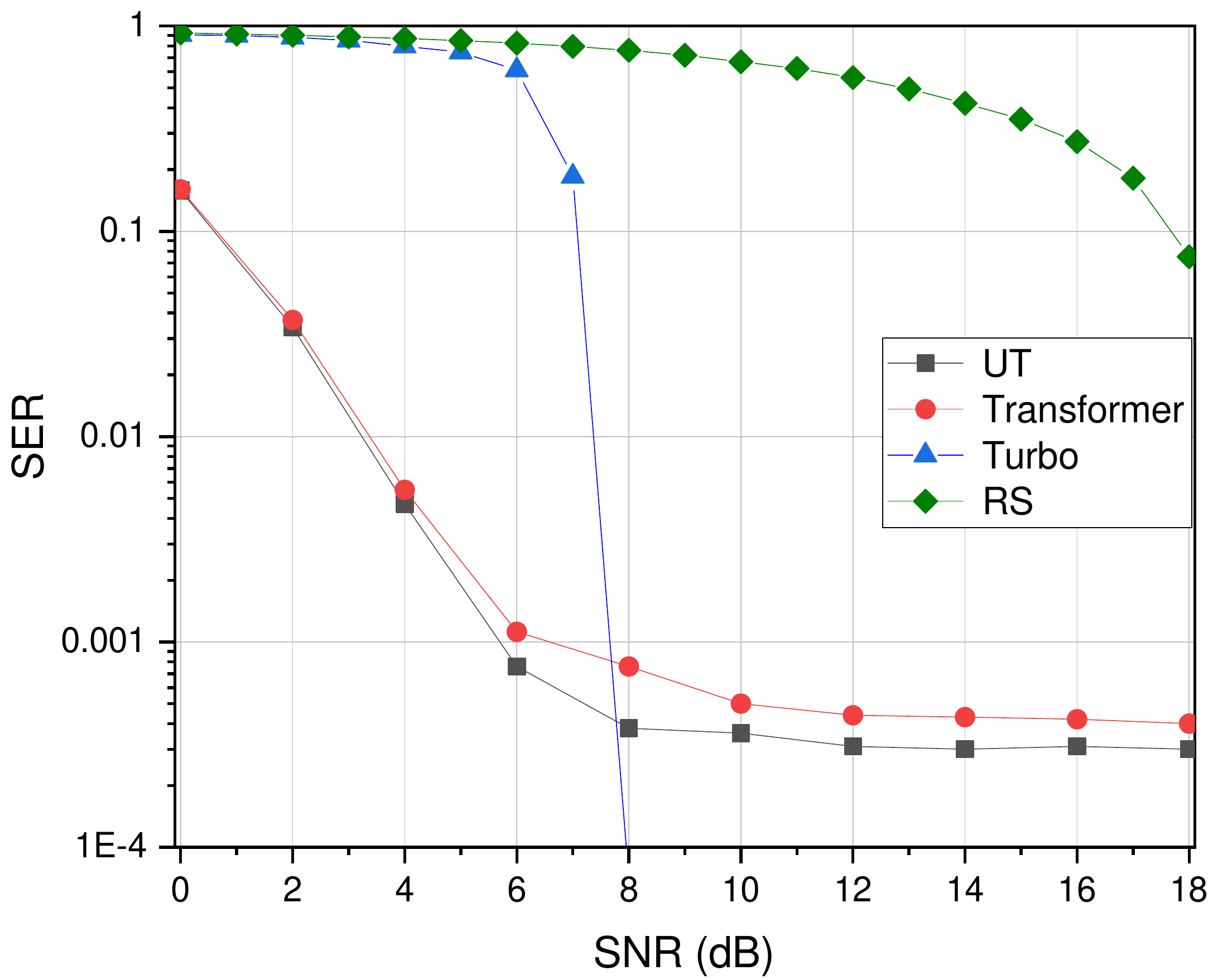}
\caption{SER score versus SNR in AWGN channel.}
\end{minipage}
\end{figure*}

\subsection{Simulation Settings}
The adopted linguistic dataset is the standard proceedings of the European Parliament [8], which consists of around 2.0 million sentences and 53 million words. The dataset is pre-processed into lengths of sentences with 4 to 30 words.  

In this numerical experiment, 
the maximum number of cycles allowed of UT, m, is set to 5, the therehold, T, is set to 0.9, learning rate for Algorithm 1 is set to $10^{-4}$, and learning rate for Algorithm 2 is set to $10^{-6}$.  The detailed settings of the proposed networks are shown in Table I.
We use the typical transformer consisted of fixed three layers for the joint design of semantic coding and channel coding as the baseline [2] [3]. The traditional approach employs source coding and channel coding schemes separately, which applies fixed-length coding (5-bit) for source coding, and Turbo coding [9] or Reed-Solomon    (RS) coding for channel coding [10]. In particular, turbo encoding rate is 1/3 and turbo decoding method is Max-Log-MAP algorithm with 5 iterations. Furthermore, the widely recognized evaluation metrics in natural language processing, namely  BLEU (bilingual evaluation understudy) [11] and Symbol Error Ratio (SER) are used to measure the performance.



\begin{table}[htb]
  \centering
  \caption{THE SETTING OF SEMANTIC COMMUNICATION SYSTEM}
\begin{tabular}{|c|c|c|c|}
\hline
 & Layer Name & Units & Activation \\ \hline
\multirow{3}{*}{\begin{tabular}[c]{@{}c@{}}Transmitter\\ (Encoder)\end{tabular}} & Embeding  & 128  & None \\ \cline{2-4} 
 & UT Encoder & 128 (8 heads) & Linear \\ \cline{2-4} 
 & Dense & 256 & ReLu \\ \cline{2-4} 
 & Dense & 16 & ReLu \\ \hline
Channel & AWGN & None & None \\ \hline
\multirow{4}{*}{\begin{tabular}[c]{@{}c@{}}Receiver\\ (Decoder)\end{tabular}} & Dense & 256 & ReLu \\ \cline{2-4} 
 & Dense & 128 & ReLu \\ \cline{2-4} 
 & UT Decoder & 128 (8 heads) & Linear \\ \cline{2-4} 
 & Predication & Dictionary & Softmax \\ \hline
\multirow{2}{*}{ACT Model} & Dense & 1 & Sigmod \\ \cline{2-4} 
 & Dense & 1 & Sigmod \\ \hline
\end{tabular}
\end{table}

\subsection{Numerical Results}
Fig. 3 and Fig. 4 show the relationship between the BLEU score and  SNR under the same number of transmitted symbols over AWGN and Rayleigh fading channel respectively, where the traditional non-semantic approaches use 64-QAM for the modulation. In Fig. 3  among the traditional baselines, Turbo coding performs better than RS coding. Obviously, Rayleigh fading channel has stronger negative influence than AWGN channel for both traditional methods and the DNN based methods. In AWGN channel, when SNR is greater than or equal to 8dB, Turbo coding can achieve almost 100$\%$ word accuracy, but RS coding performs much worse than Turbo coding at the same SNR. All the DNN based approaches (i.e., transformer and UT) achieve better results than the traditional ones. In particular, the semantic communication based on UT obtains a better word accuracy than the semantic communication with transformer. We can also observe that when SNR = 4 dB, both the semantic communication with either UT or transformer can achieve almost 100$\%$ word accuracy in terms of BLEU score under the AWGN channel (Fig 3(a)), which implies it is hard for us to use BLEU to distinguish the performance of the two approaches. So it’s helpful to further introduce  SER to make accurate distinction between the proposed semantic communication with UT and the existing  semantic communication with transformer. 

Fig. 5 demonstrates the relationship between the SER and SNR. As we can see from the curves in Fig. 5, the semantic communication system with UT shows a better performance than the semantic communication system with transformer. The introduction of adaptive circulation mechanism in the semantic communication system make the whole system more flexible to deal with the situation of varying physical channel with different SNR, being adaptve to different SNR to make transmission adjustments, so as to improve the communication performance. We can also observe that there exists an error floor for all the transformer based mechanisms, but this phenomenon don’t happen  in traditional approaches. We notice that neural networks based solutions are suboptimal in high SNR region as mentioned in [14]. Since this kind of error is hard to encounter in conventional neural networks (with probability less than $10^{-4}$), it is really hard to get suitable examples to train the neural network. Thus semantic encoding \& decoding mechanism in the region of high SNR shows diminishing effects, which generates an avoidable error floor for the semantic communication system. How to improve semantic coding performance in high SNR region with data imbalance is an interesting theme for future research direction.



\begin{figure*}[ht]
\centering
\begin{minipage}{0.24\textwidth}
\includegraphics[width=4.2cm,height=5cm]{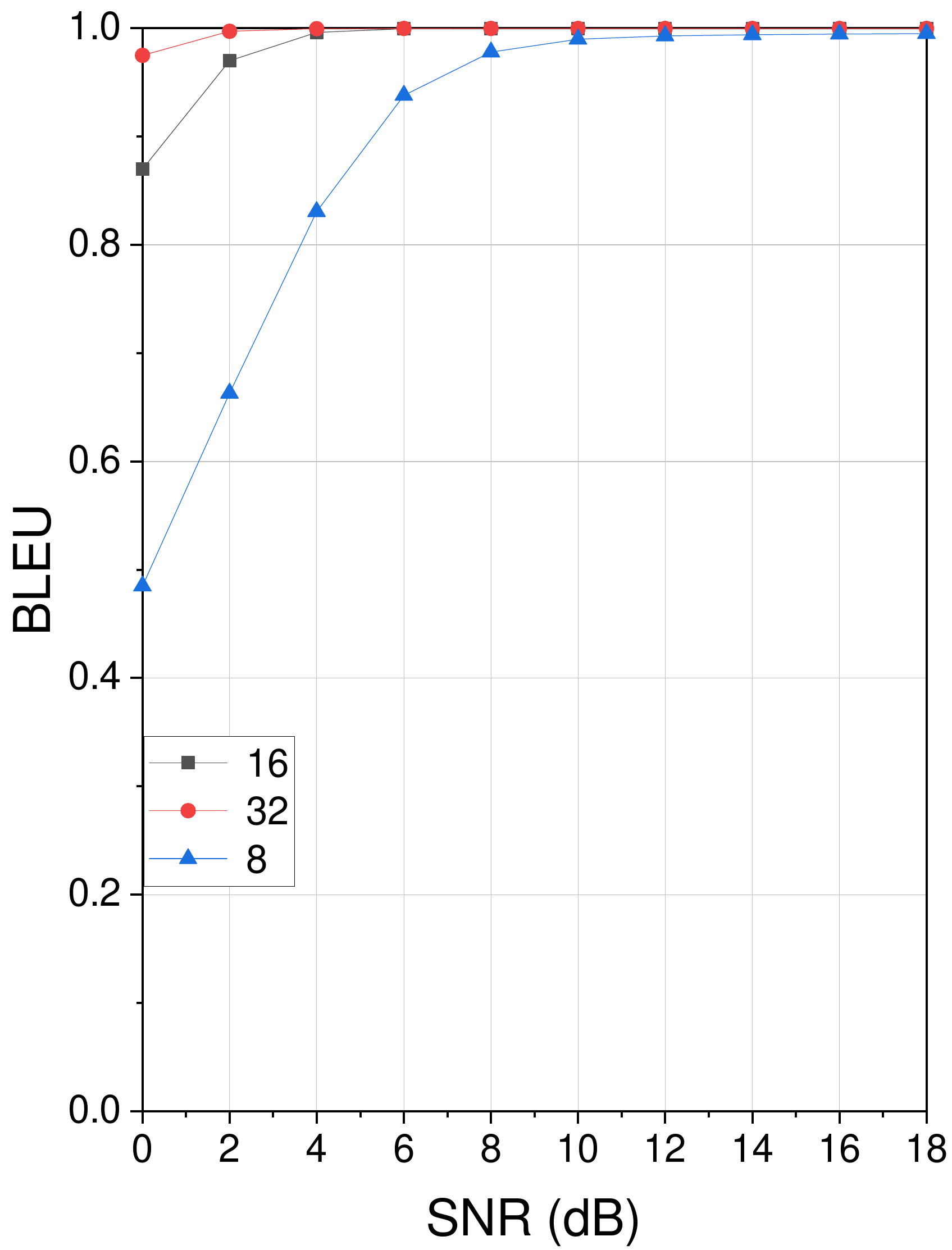}
\caption{BLEU versus the number of symbols used for in one word in semantic communication.}
\end{minipage}
\begin{minipage}{0.24\textwidth}
\includegraphics[width=4.2cm,height=5cm]{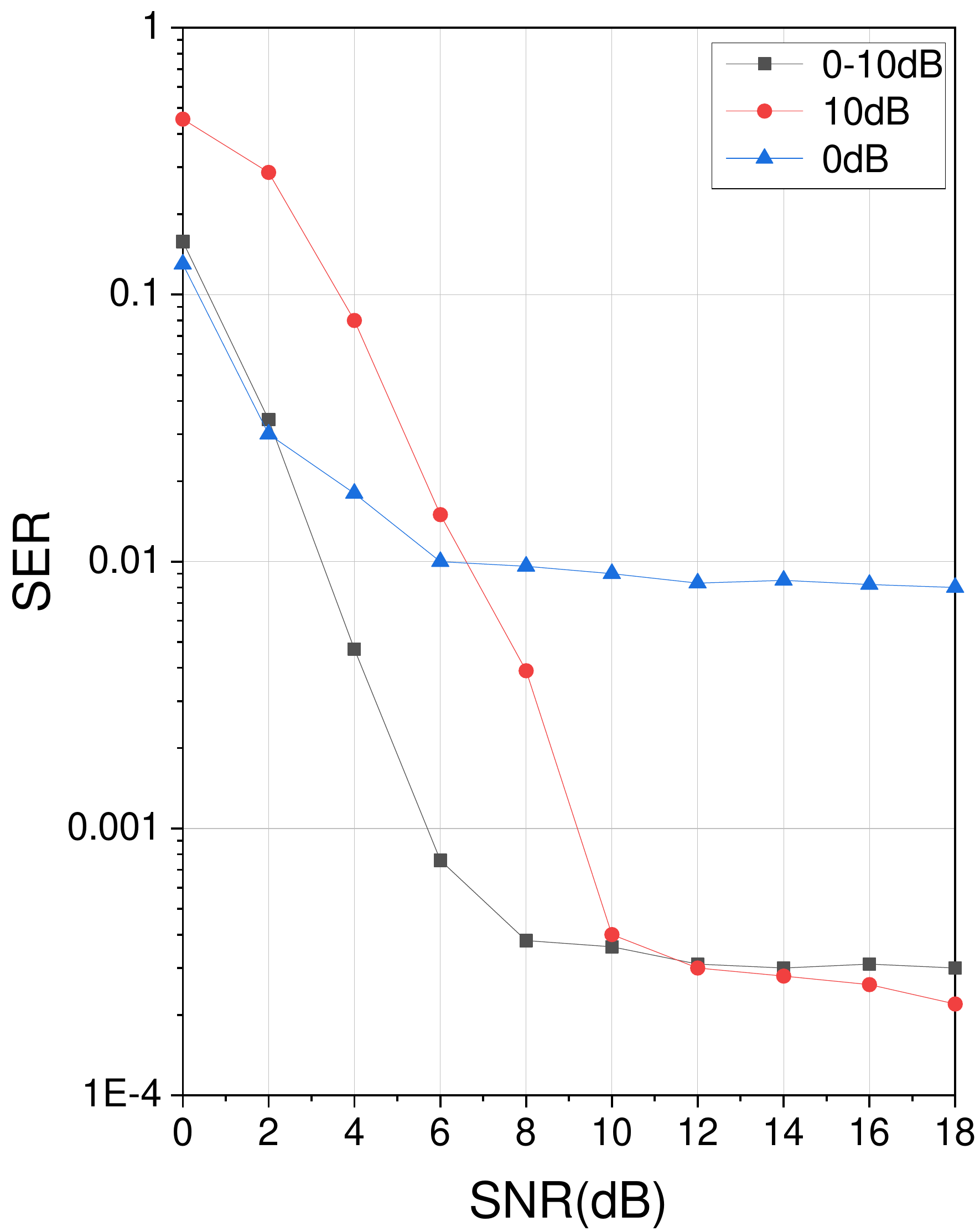}
\caption{The impact of different training SNR for semantic communication system.}
\end{minipage}
\begin{minipage}{0.24\textwidth}
\includegraphics[width=4.2cm,height=5cm]{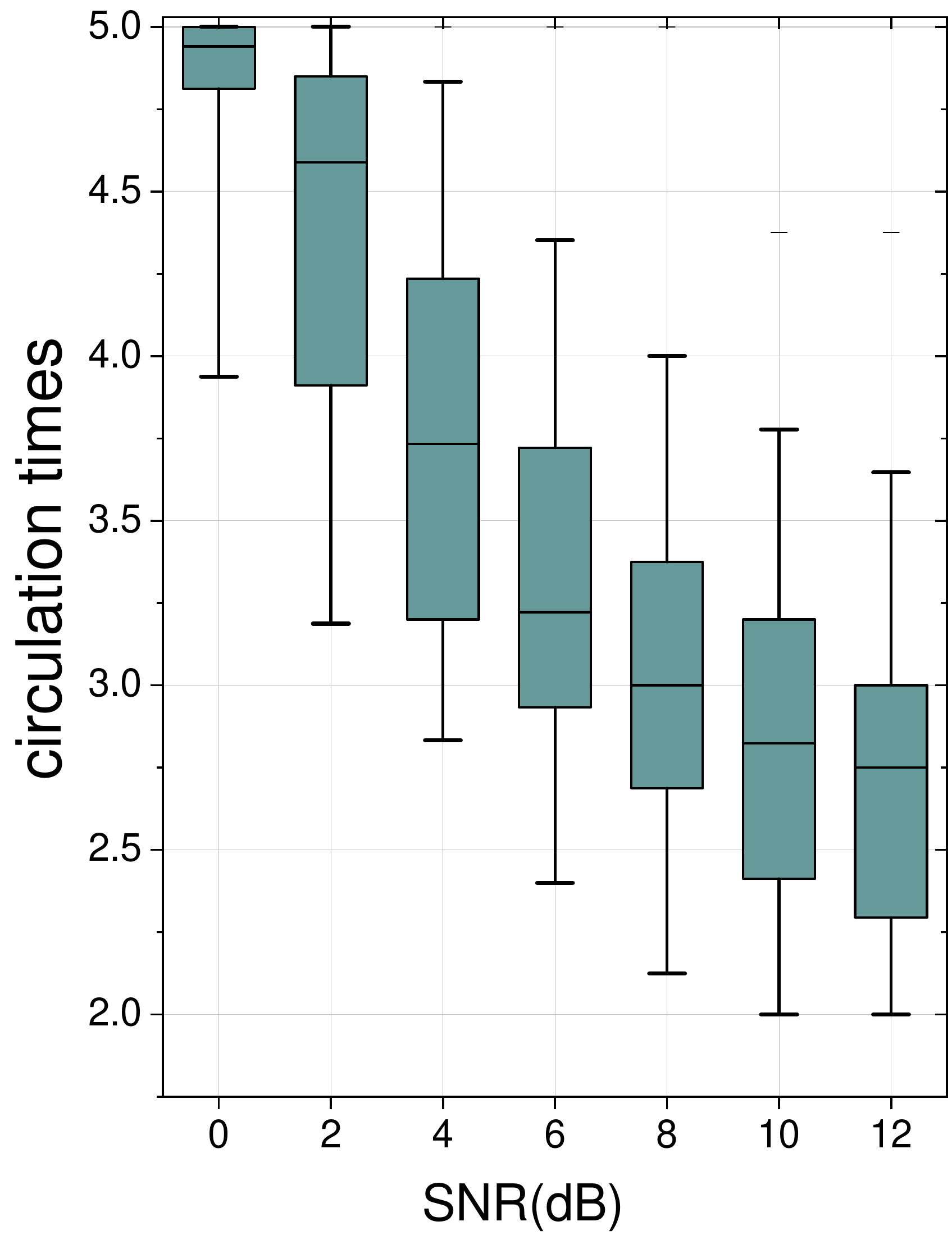}
\caption{The relationship between computational complexity and SNR for semantic communication system.}

\end{minipage}
\begin{minipage}{0.24\textwidth}
\includegraphics[width=4.2cm,height=5cm]{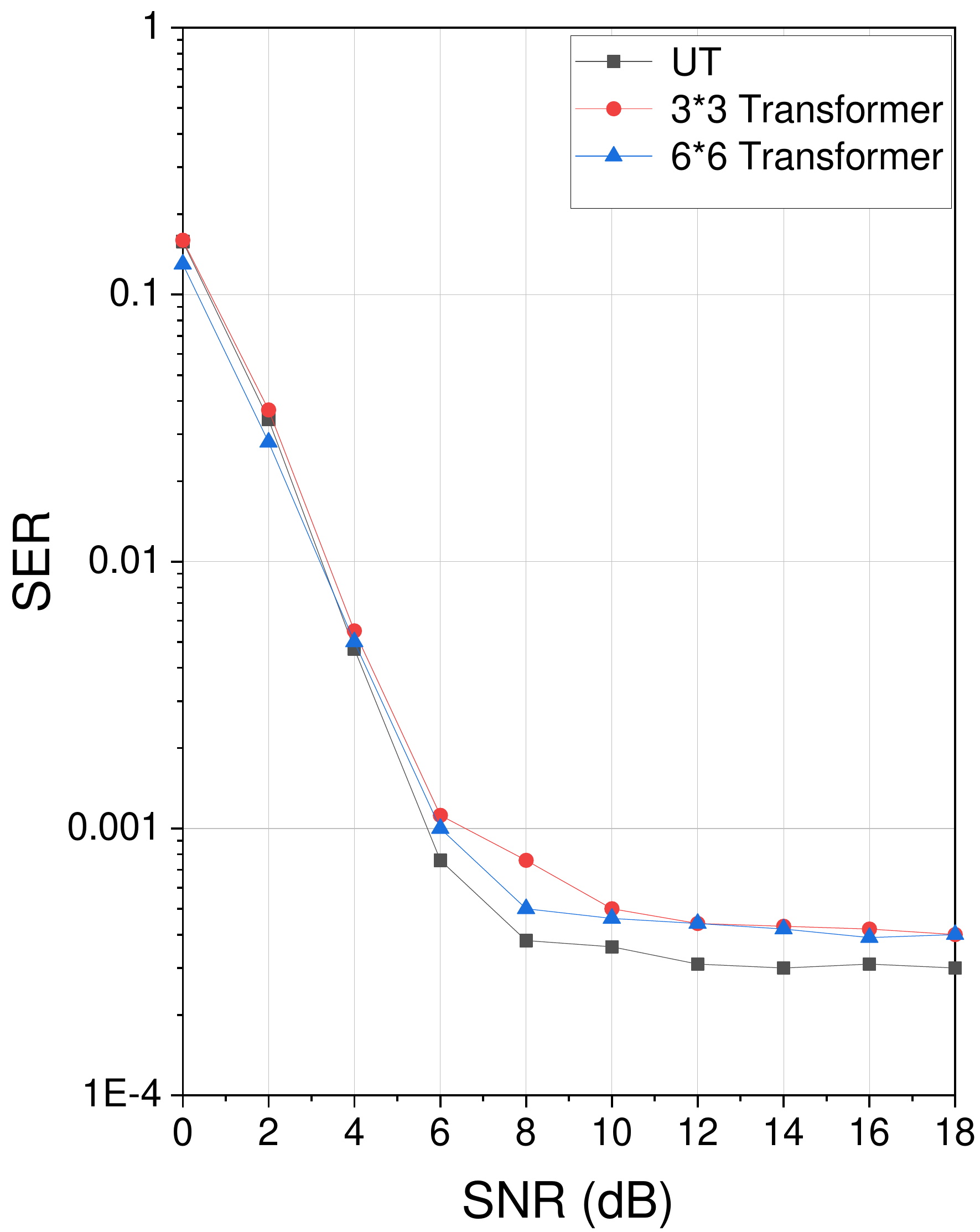}
\caption{The semantic communication system based on DNNs with different layers.}
\end{minipage}
\end{figure*}






Fig. 6 displays the impact of the number of symbols per word on the BLEU score and the SER respectively. As the number of symbols per word grows, the BLEU score and SER increase significantly due to the increasing distance between constellations. The greater the number of symbols per word means the more noise-tolerant is each word. We can also observe that the improvement becomes marginal with the increasing of symbols per word. It means that we can use fewer symbols to represent one word in the process of transmission with relatively small performance loss. On the other hand, one way to increase the BLEU score in low SNR region is to increase the number of symbols for each word.


Noise level in training is another critical parameter for training the transformer and the effects can be seen in the Fig. 7. When  training is carried out in high SNR region with high noise level (i.e.10 dB),   UT performs better even the noise level is high, but it can not correctly transmit the sentence in low SNR region. When training is carried out in low SNR region (i.e. 0dB), UT performs better since the noise level is low, but when SNR increase to the high region, it can not transmit the sentence as similarly accurate as UT trained in the previous case. Hence , we choose to train UT in a manner with SNR changing randomly between 0-10dB, which ensures the accuracy of information transmission meanwhile guarantees the ability to eliminate strong noise.



Fig. 8 shows the fluctuation of the number of cycles required by the whole system in the face of sentences with different complexity in different SNR. In order to meet the preset threshold for the confidence of the final output sentences, UT will increase the number of cycles of sentences when faced with more complex sentences, so as to ensure the accurate transmission of information. Similarly, when the channel under low SNR, the transmitted information will be seriously disturbed by noise. In the case of the same cycle times, the confidence of the information can no longer meet the preset threshold, so UT will correspondingly increase the cycle times of the information in the semantic layer during information processing, so as to ensure the accuracy of the transmission.


Besides  using UT to improve the performance of semantic communication system, we can consider to increase the attention layers for the original structure of the conventional transformer. As illustrated in Fig. 9, the conventional transformer with increased layers (i.e. transformer with 6*6 layers) performs a little bit better in low SNR region, however in high SNR region, there is almost no performance improvement. It may be due to the reason that when in low SNR region, using the lower number of layers (i.e. transformer with 3*3 layers) is enough to eliminate the noise influence. Redundant layers will not improve the performance of the system, and even the whole system may have over-fitting phenomenon because of the increased parameter complexity. Surely, the approach of increasing the layers improve the system performance at the cost of increasing the number of the parameters, by which the difficulty of convergence is increased as well. On the contrary, through the using of UT, we can improve the performance of the system without increasing the number of parameters, meanwhile the system can better adapt to a variety of changing channel conditions. This can be a promising way to improve the performance of semantic communication system.

\section{Conclusion}
This Letter proposed a new semantic communication system based on adaptive Universal Transformer. By combining the circulation mechanism with transformer, the semantic communication system becomes more flexible and powerful, and can better adapt to various channel conditions. It is the first attempt to adopt Universal Transformer for the semantic communication system, responsive to the difference of semantic information contained in different sentences. Although there exists a SER error floor and the corresponding improvement becomes marginal when the level of SNR exceeds certain value, we believe that our results can be a promising start for paving the way towards the era of  semantic communications.

\bibliographystyle{IEEEtran}
\bibliography{IEEEabrv,reference. bib}
\nocite{*}
\end{document}